\documentclass[letterpaper]{article} % DO NOT CHANGE THIS
\usepackage[table]{xcolor} 
\usepackage{main}  % DO NOT CHANGE THIS
\usepackage{times}  % DO NOT CHANGE THIS
\usepackage{helvet}  % DO NOT CHANGE THIS
\usepackage{courier}  % DO NOT CHANGE THIS
\usepackage[hyphens]{url}  % DO NOT CHANGE THIS
\usepackage{graphicx} % DO NOT CHANGE THIS
\usepackage{natbib}  % DO NOT CHANGE THIS AND DO NOT ADD ANY OPTIONS TO IT
\usepackage{caption} % DO NOT CHANGE THIS AND DO NOT ADD ANY OPTIONS TO IT
\usepackage{algorithm}
\usepackage{algorithmic}

\usepackage{newfloat}
\usepackage{listings}
\DeclareCaptionStyle{ruled}{labelfont=normalfont,labelsep=colon,strut=off} % DO NOT CHANGE THIS
\floatstyle{ruled}
\newfloat{listing}{tb}{lst}{}
\floatname{listing}{Listing}

\usepackage{enumitem}
\usepackage{amsmath}
\usepackage{amssymb}
\usepackage{tcolorbox}
\usepackage{multirow}
\usepackage{booktabs}
\usepackage[table]{xcolor}
\usepackage{colortbl}
\usepackage{tabularx}
\usepackage{hyperref}
\usepackage{flushend}

\definecolor{lightgold}{rgb}{1.0, 0.95, 0.8}

\author{
    Yuhang Liu\textsuperscript{\rm 1,3}\equalcontrib\thanks{Work done during an internship at InfiX.ai.},
    Zeyu Liu\textsuperscript{\rm 2}\equalcontrib,
    Shuanghe Zhu\textsuperscript{\rm 1},
    Pengxiang Li\textsuperscript{\rm 2},
    Congkai Xie\textsuperscript{\rm 3},\\
    Jiasheng Wang\textsuperscript{\rm 4,3}\footnotemark[2],
    Xavier Hu\textsuperscript{\rm 1},
    Xiaotian Han,
    Jianbo Yuan\textsuperscript{\rm 5}\thanks{Work done outside of Amazon.},
    Xinyao Wang\textsuperscript{\rm 5}\footnotemark[3],\\
    Shengyu Zhang\textsuperscript{\rm 1}\thanks{Corresponding author.},
    Hongxia Yang\textsuperscript{\rm 2,3}\footnotemark[4],
    Fei Wu\textsuperscript{\rm 1}
}
\affiliations{
    \textsuperscript{\rm 1}Zhejiang University
    \textsuperscript{\rm 2}The Hong Kong Polytechnic University
    \textsuperscript{\rm 3}InfiX.ai
    \textsuperscript{\rm 4}The University of Chicago
    \textsuperscript{\rm 5}Amazon \\
    \texttt{\{liuyuhang, sy\_zhang\}@zju.edu.cn},
    \texttt{hongxia.yang@polyu.edu.hk}
}

\nocopyright
\usepackage{bibentry}
\begin{document}

\title{
    \centering
    \begin{tabular}{c c}
      \multirow{2}{*}{\includegraphics[width=1.4cm]{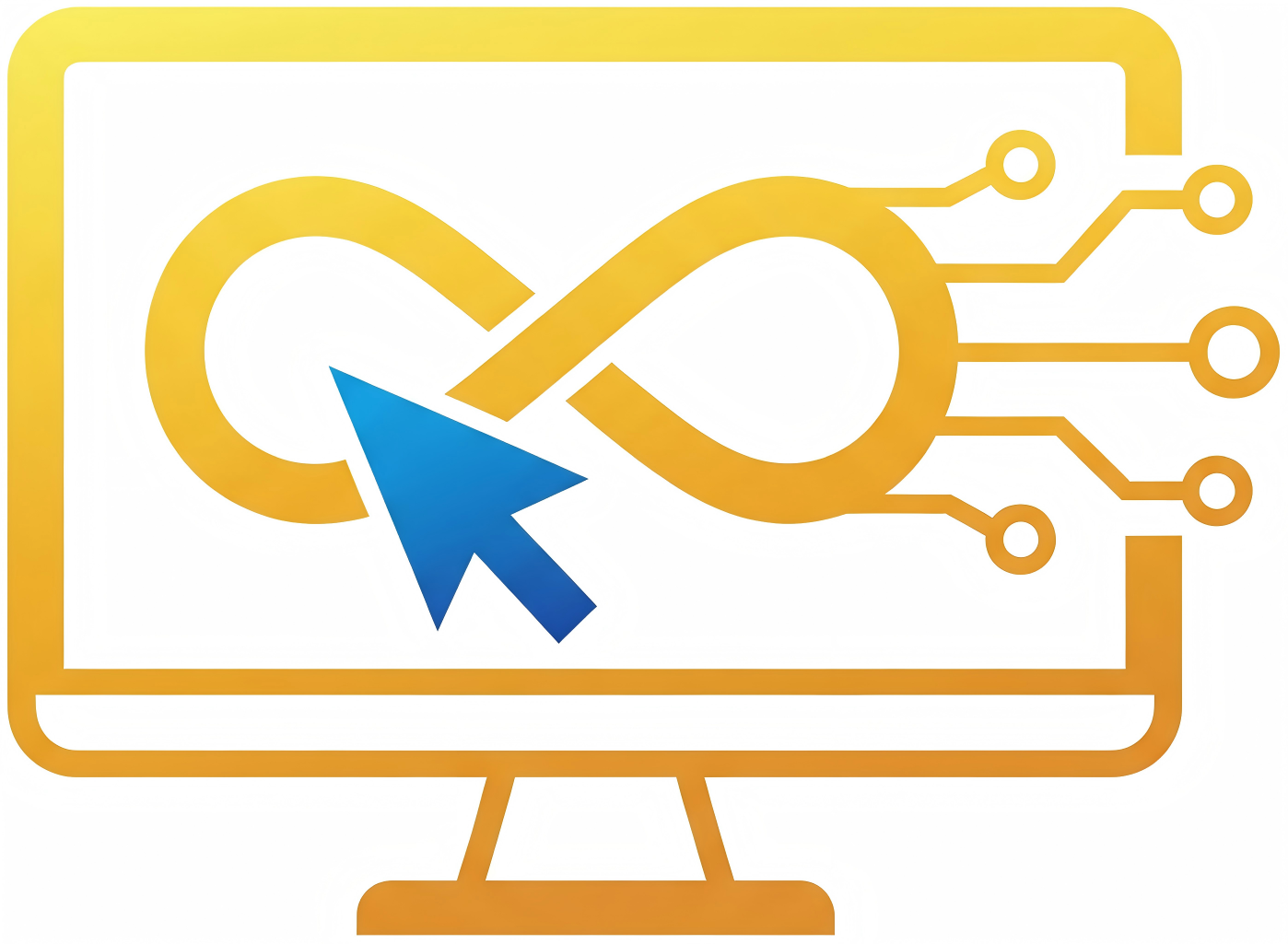}}
      & InfiGUI-G1: Advancing GUI Grounding with Adaptive \\
      & Exploration Policy Optimization
    \end{tabular}
}

\maketitle

\begin{abstract}
The emergence of Multimodal Large Language Models (MLLMs) has propelled the development of autonomous agents that operate on Graphical User Interfaces (GUIs) using pure visual input. A fundamental challenge is robustly grounding natural language instructions. This requires a precise \textit{ spatial alignment}, which accurately locates the coordinates of each element, and, more critically, a correct \textit{ semantic alignment}, which matches the instructions to the functionally appropriate UI element. Although Reinforcement Learning with Verifiable Rewards (RLVR) has proven to be effective at improving \textit{spatial alignment} for these MLLMs, we find that inefficient exploration bottlenecks \textit{semantic alignment}, which prevent models from learning difficult semantic associations. To address this exploration problem, we present Adaptive Exploration Policy Optimization (AEPO), a new policy optimization framework. AEPO employs a multi-answer generation strategy to enforce broader exploration, which is then guided by a theoretically grounded Adaptive Exploration Reward (AER) function derived from first principles of efficiency $\eta=U/C$. Our AEPO-trained models, InfiGUI-G1-3B and InfiGUI-G1-7B, establish new state-of-the-art results across multiple challenging GUI grounding benchmarks, achieving significant relative improvements of up to 9.0\% against the naive RLVR baseline on benchmarks designed to test generalization and semantic understanding. Resources are available at \urlstyle{tt}\url{https://github.com/InfiXAI/InfiGUI-G1}.
\end{abstract}

\section{Introduction}
\label{sec:introduction}

\begin{figure}[t]
\centering
\includegraphics[width=0.9\linewidth]{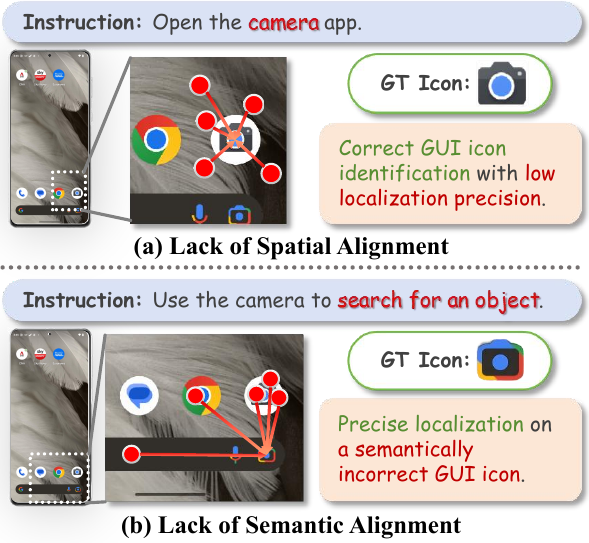}
\caption{Primary GUI‐grounding failure modes. 
(a) \textbf{Spatial‐alignment} failure: the model selects the correct icon but localizes it imprecisely. (b) \textbf{Semantic‐alignment} failure: the model localizes precisely on an incorrect icon due to misinterpreting the instruction. Although RLVR methods have advanced spatial alignment, semantic alignment remains the critical bottleneck for complex GUI tasks—this work is devoted to addressing it.}
\label{fig:introduction}
\end{figure}

The development of autonomous agents capable of operating across the vast landscape of graphical user interfaces (GUIs) is a key frontier in achieving general-purpose human-computer interaction \cite{wang2024ponderpressadvancing}. The success of these agents is fundamentally predicated on a core perceptual task: GUI Grounding. This task involves accurately mapping a natural language instruction to a specific interactive element on a screen. 
The challenge of GUI Grounding can be deconstructed into two orthogonal dimensions: \textit{Spatial Alignment}, which focuses on the precision of locating an element (i.e., "pointing" accurately), as shown in Fig.~\ref{fig:introduction}(a). \textit{Semantic Alignment}, which pertains to the correctness of identifying the appropriate element to interact with (i.e., "pointing" at the right target), as illustrated in Fig.~\ref{fig:introduction}(b). Robust and reliable agent performance in complex, real-world scenarios hinges on proficiency in both, with Semantic Alignment being particularly critical.

Current fine-tuning methodologies for multimodal large language models (MLLMs) face major challenges in achieving robust \textit{spatial alignment} and \textit{semantic alignment}. While Supervised Fine-Tuning (SFT) can be effective, it is highly data-intensive and struggles to generalize to unseen UI layouts \cite{cheng2024seeclick}. By contrast, Reinforcement Learning with Verifiable Rewards (RLVR) improves data efficiency by optimizing sequential coordinate generation, which has proven effective at enhancing spatial alignment \cite{yuan2025enhancingvisualgroundinggui}. 

However, most of existing RLVR methods share one limitation: \textbf{inefficient exploration}. They rely on the model's current policy to sample actions and thus get stuck on high-confidence errors. This ``confidence trap" prevents discovery of low-probability but correct actions, bottlenecking \textit{semantic alignment}. 
As shown in Fig.~\ref{fig:introduction}(b), when the instruction is ``Use the camera to search for an object" on a screen displaying various icons, a model with weak semantic understanding may repeatedly select the generic ``Camera" button. Standard RLVR would keep sampling this high-confidence but incorrect ``Camera'' icon, rarely stumbling upon the correct ``Google Lens" icon, and thus fail to receive the learning signal necessary to correct its semantic misunderstanding.

We introduce \textbf{Adaptive Exploration Policy Optimization (AEPO)}, a novel approach to overcome the exploration bottleneck in standard RL. By integrating the \textbf{multi-answer generation strategy}, AEPO drives the model to explore a diverse set of candidate solutions in a single forward pass, addressing the limitations of standard RL, which struggles with low sampling efficiency and the strategy confidence trap. Complemented through the \textbf{adaptive exploration reward (AER)}, a non-linear reward signal, AEPO dynamically guides exploration, promoting exploration during failures and convergence upon successes, while avoiding the simplistic or distance-based rewards. Additionally, the \textbf{quality-of-exploration penalty} ensures high-quality exploration by penalizing inefficient, near-collinear outputs, fostering true semantic diversity rather than simplistic linear scans in the geometric space. In summary, the key contributions of our work are as follows:
\begin{itemize}
    \item We present a novel policy-optimization method, \textbf{Adaptive Exploration Policy Optimization (AEPO)}, which integrates multi-answer generation into the reinforcement learning framework to boost exploration efficiency for GUI grounding significantly.
    \item To balance the trade-off between exploration and exploitation,  we devise an \textbf{Adaptive Exploration Reward (AER)} that incentivizes models to explore both extensively and purposefully.
    \item Building on the above framework, we introduce the \textbf{InfiGUI-G1} series model—3B and 7B variants—whose extensive evaluation across diverse benchmarks establishes a state-of-the-art in the GUI grounding task.
\end{itemize}

\section{Related Work}
\label{sec:related_work}

\subsection{MLLM-based GUI Agents and Grounding}
\label{sec:related_gui_agents}

Recently, the paradigm for GUI automation has shifted gradually from brittle, script-based methods to visually driven, human-like approaches. A representative early attempt, OmniParser \cite{lu2024omniparserpurevisionbased}, utilizes an MLLM (e.g., GPT-4V \cite{yang2023dawnlmmspreliminaryexplorations}) to parse visual UI elements in a screenshot into traditional structured data. 
OS-Atlas \cite{wu2024osatlasfoundationactionmodel} and U-Ground \cite{gou2025navigatingdigitalworldhumans} explored hybrid interfaces, intending to achieve robust and flexible performance across diverse environments~\cite{nguyen2024guiagentssurvey}. Notably, SeeClick \cite{cheng2024seeclick} firstly completed GUI tasks via relying solely on screenshots (visual input) and MLLMs, promising greater adaptability and cross-platform universality.  
However, its approach introduced a new task—GUI grounding—which has been identified as a key metric in this paradigm but also as a primary performance bottleneck.

To address GUI grounding, researchers have advanced a spectrum of techniques that enhance MLLMs’ visual-locating capabilities. These include large-scale pre-training on GUI-specific corpora \cite{qin2025uitarspioneeringautomatedgui, yang2025magmafoundationmodelmultimodal, wu2025guireflectionempoweringmultimodalgui}, targeted supervised fine-tuning (SFT) \cite{yang2025ariauivisualgroundinggui, hui2025winclickguigroundingmultimodal}, and reasoning-oriented frameworks \cite{luo2025guir1generalistr1style, lee2025reguidedataefficientgui, wei2025learningreasoningrefinementframework}. In parallel, novel training techniques have been adapted for MLLMs, including coordinate-free methods that generate attention maps instead of explicit coordinates~\cite{wu2025guiactorcoordinatefreevisualgrounding}, and inference-time optimization strategies that elevate performance without retraining~\cite{wu2025dimoguiadvancingtesttimescaling}.

\subsection{Reinforcement Learning in MLLM}
\label{sec:related_rlmllm}

Reinforcement learning has rapidly become a potent paradigm for sharpening the reasoning capabilities of multimodal large language models. Building on the recent success of DeepSeek-R1~\cite{deepseekai2025deepseekr1incentivizingreasoningcapability} in large language models, a succession of vision-centric models, such as Vision-R1 \cite{huang2025visionr1incentivizingreasoningcapability}, Visual-RFT \cite{liu2025visualrftvisualreinforcementfinetuning}, MedVLM-R1 \cite{pan2025medvlmr1incentivizingmedicalreasoning}, InfiMMR~\cite{liu2025infimmr}, demonstrated RL’s broad potential across diverse domains~\cite {zhou2025reinforcedmllmsurveyrlbased}.

In the context of GUI grounding, RL has demonstrated practical applicability through several notable approaches \citep{liu2025infigui,zhou2025gui,tang2025guig2gaussianrewardmodeling,lian2025ui,yang2025gta1}. UI-R1 \cite{lu2025uir1enhancingefficientaction} introduces a novel rule-based action reward mechanism that enables model optimization using policy-based algorithms. GUI-R1 \cite{luo2025guir1generalistr1style} adopts a unified action space modeling strategy, which extracts and integrates action space categories across different platforms into a cohesive framework.
Additionally, self-supervised \cite{gao2025uishiftenhancingvlmbasedgui} and self-evolutionary \cite{yuan2025enhancingvisualgroundinggui} RL methods have been proposed to address the limitations of traditional supervised fine-tuning (SFT), which often relies on large amounts of diverse labeled data. Reinforcement fine-tuning \cite{zhang2025agentcpmguibuildingmobileuseagents} also shows promise as a pathway toward integrated training.
R-VLM \cite{park2025rvlmregionawarevisionlanguage} introduces a two-stage zoom-in grounding process that refines predictions through a zoomed-in view of region proposals. This is combined with an IoU-aware weighted cross-entropy loss to enhance fine-grained perception in grounding tasks.
Overall, RL has proven to be an effective and efficient approach for training multi-modal large language models (MLLMs) and advancing GUI grounding performance.

Notably, these methods are constrained by a single-answer generation paradigm, which leads to inefficient exploration and can reinforce the model's confident but incorrect behaviors. In contrast, our framework employs multi-answer generation to enforce a broader search, which is then guided by our adaptive exploration reward function to provide richer and more effective learning signals.

\section{Methodology}
\label{sec:methods}
This section details our proposed AEPO framework. We first formalize the GUI grounding task as a policy optimization problem in \S\ref{sec:problem_formulation}. We then elaborate on the core components of the AEPO framework in \S\ref{sec:aepo}, including multi-answer generation (\S\ref{sec:multi_answer}), the adaptive exploration reward (\S\ref{sec:aer}), and the collinear penalty (\S\ref{sec:collinear_penalty}). Finally, we present the overall training objective in \S\ref{sec:overall_objective}.

\subsection{Problem Formulation}
\label{sec:problem_formulation}
We formulate GUI grounding as a direct policy optimization problem. The goal is to train a policy $\pi_\theta$, represented by an MLLM with parameters $\theta$, to generate an action that correctly corresponds to a given context.
\begin{itemize}
    \item \textbf{Context} $c$: A tuple $(\mathcal{S}, \mathcal{I})$, where $\mathcal{S}$ is a GUI screenshot and $\mathcal{I}$ is a natural language instruction.
    \item \textbf{Action} $a$: The output generated by the policy, which is a coordinate point $p=(x,y)$.
    \item \textbf{Ground Truth} $B$: The ground truth bounding box of the target UI element corresponding to the instruction $\mathcal{I}$.
    \item \textbf{Policy} $\pi_\theta(a|c)$: The policy defines the probability distribution over all possible actions given a context $c$.
    \item \textbf{Reward Function} $R(a, B)$: A deterministic function that returns a scalar reward. For a generated point $p$, the reward is positive if $p \in B$ and negative otherwise.
\end{itemize}
The objective is to find the optimal parameters $\theta^*$ that maximize the expected reward over the data distribution $\mathcal{D}$:
\begin{equation}
    \theta^* = \arg\max_\theta \mathbb{E}_{c \sim \mathcal{D}, a \sim \pi_\theta(\cdot|c)} [R(a, B)]
\end{equation}
Because the action $a$ (i.e., the coordinate string) is generated auto-regressively, its sequential generation process is well-suited for optimization with policy gradient algorithms from reinforcement learning, such as Proximal Policy Optimization (PPO, \citet{schulman2017proximal}), Group Relative Policy Optimization (GRPO, \citet{shao2024deepseekmath}), or REINFORCE Leave-One-Out (RLOO, \citet{ahmadian2024back}).

\subsection{Adaptive Exploration Policy Optimization}
\label{sec:aepo}
\begin{figure*}[!tp]
\centering
\includegraphics[width=\textwidth]{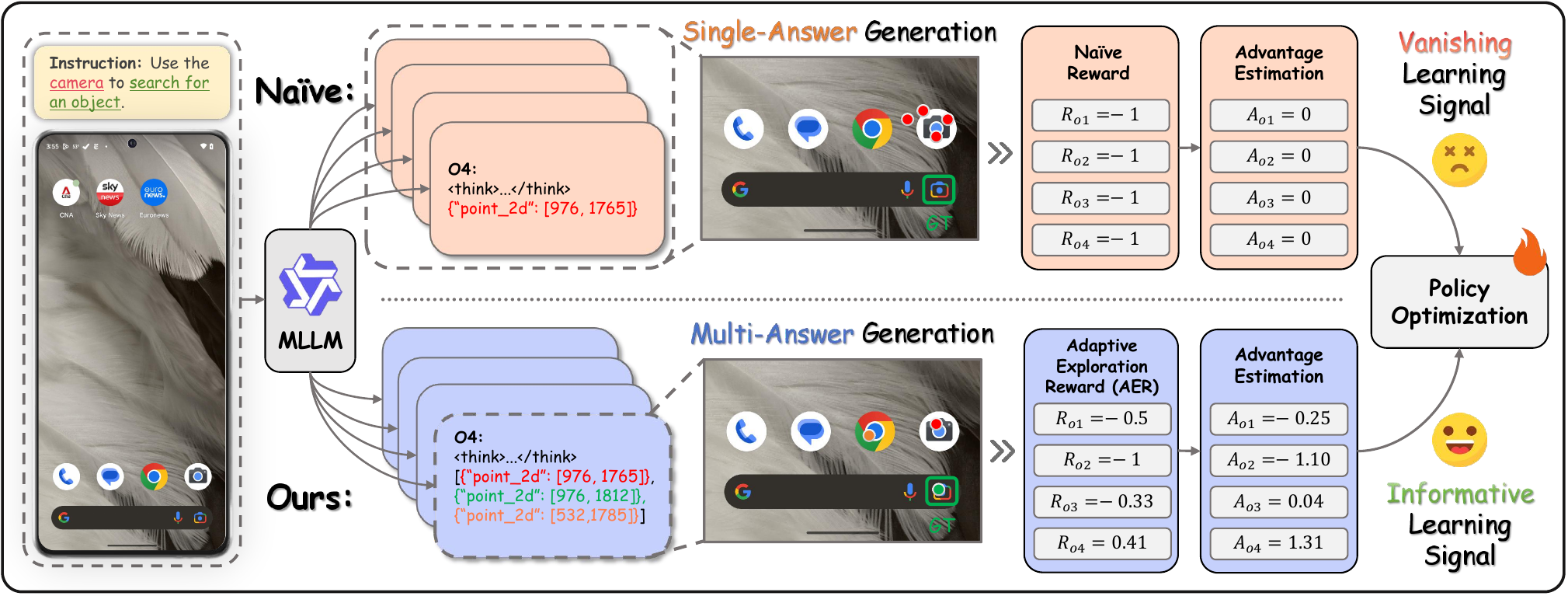}
\caption{Comparison of AEPO and a naive RL baseline.
\textbf{Top:} The naive single-answer approach becomes trapped on high-confidence errors, repeatedly sampling the same incorrect action and producing a \textit{vanishing learning signal} when no positive reward is discovered.
\textbf{Bottom:} \textbf{AEPO} employs multi-answer generation to explore diverse candidates each rollout and an \textbf{AER} to derive an \textit{informative learning signal} from their efficiency and correctness. These mechanisms break the exploration bottleneck in GUI agents and enable robust semantic alignment.
}
\label{fig:method}
\end{figure*}

To overcome the exploration limitations of the standard formulation, we introduce a novel framework, namely Adaptive Exploration Policy Optimization (AEPO), as depicted in Fig.~\ref{fig:method}. AEPO enhances the policy optimization process through three synergistic components. The \textbf{multi-answer generation} mechanism enhances RL by improving exploration of suboptimal correct answers, overcoming low sampling efficiency and the strategy confidence trap. The \textbf{adaptive reward function} fosters exploration in response to failure while driving convergence upon success. The \textbf{quality-of-exploration penalty} improves exploration quality, ensuring that "multi-answer generation" promotes true diversity in the semantic space, beyond a mere linear scan in the geometric space.

\paragraph{Multi-Answer Generation.}
\label{sec:multi_answer}
To fundamentally bypass the exploration bottleneck, our mechanism prompts the model to generate a set of $N$ candidate points, $\mathcal{A} = \{p_1, p_2, ..., p_N\}$, in a single forward pass. This forces the model to look beyond its single most confident prediction, significantly increasing the probability of sampling a correct action from the tail of the policy's distribution, especially for semantically challenging samples.

\paragraph{Adaptive Exploration Reward.}
\label{sec:aer}
\begin{figure}[!tp]
\centering
\includegraphics[width=\linewidth]{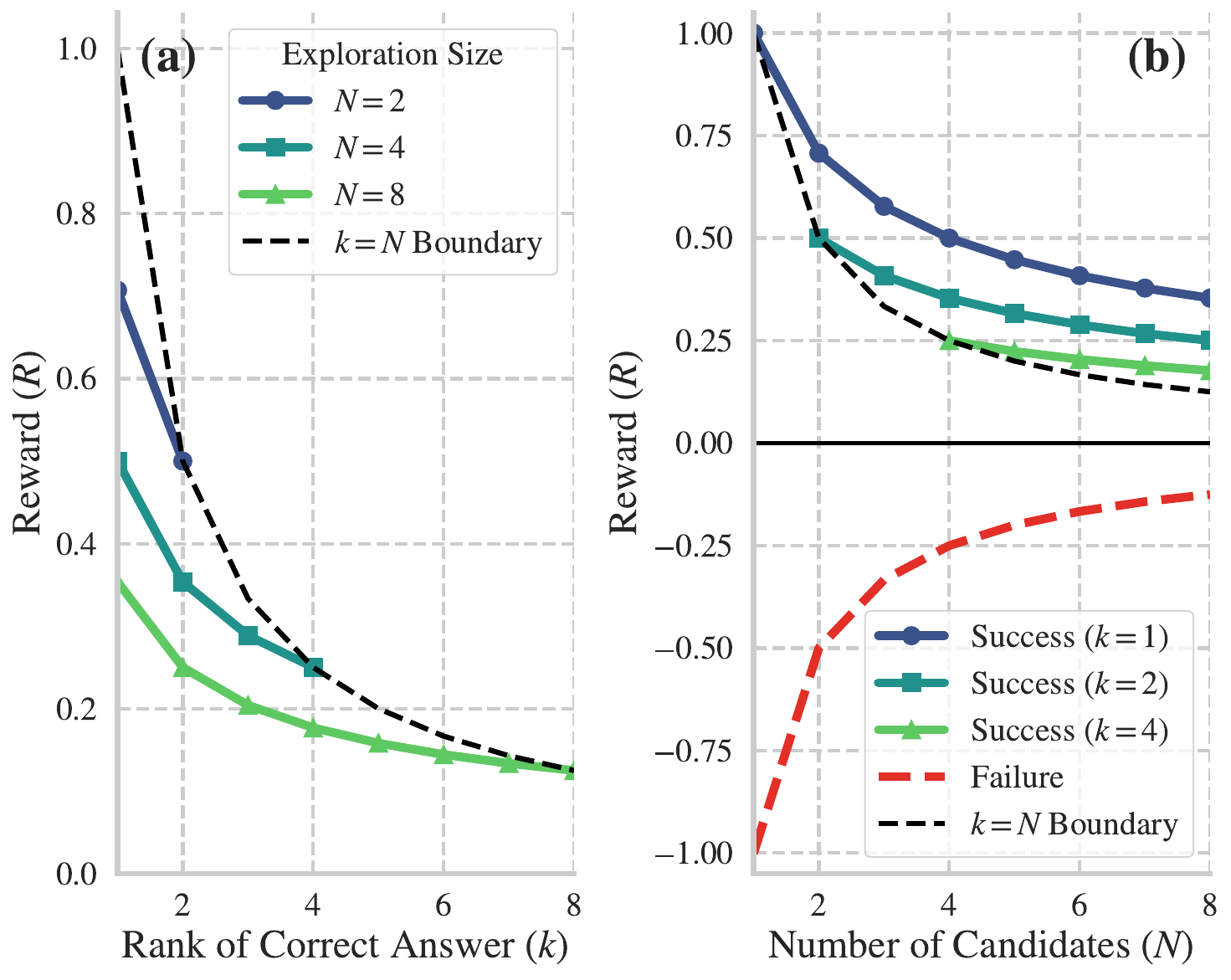}
\caption{
Visualization of the AER function based on the efficiency ratio $\eta = U/C$. 
\textbf{(a)} The reward curve increases nonlinearly to strongly incentivize selection of the correct answer, i.e., lower rank $k$. 
\textbf{(b)} The AER dynamically balances exploration and exploitation: successful trials (green/blue curves) receive higher reward for greater efficiency (smaller candidate set $N$), whereas failures (red curve) incur diminishing penalties to promote broader exploration.
}
\label{fig:reward_curve}
\end{figure}

AER provides an adaptive reward signal to guide the multi-answer exploration process. It is derived from a first-principles model of efficiency, $\eta = U/C$, where $U$ is utility and $C$ is cost.
\begin{itemize}
    \item \textbf{Utility ($U$):} The utility is defined by the outcome of the exploration. If any point $p_i \in \mathcal{A}$ falls within the ground truth bounding box $B$, the exploration is a success ($U=+1$). Otherwise, it is a failure ($U=-1$), reflecting not only the wasted computational resources but also the risk of guiding the agent into an erroneous state.
    \item \textbf{Cost ($C$):} The cost is modeled as the geometric mean of two components. The \textbf{proposal cost}, $C_p=N$, represents the effort to generate $N$ candidates. The \textbf{verification cost}, $C_v$, represents the subsequent effort to identify the correct answer. We use the geometric mean, $C = \sqrt{C_p \cdot C_v}$, as it appropriately captures the diminishing marginal returns of improving an already high-ranked answer. In case of success, $C_v=k$ (the rank of the first correct point), leading to $C_{\text{success}} = \sqrt{N \cdot k}$. In case of failure, all $N$ points must be checked, so $C_v=N$, and $C_{\text{failure}} = \sqrt{N \cdot N} = N$.
\end{itemize}
This leads to the AER function, which defines the accuracy component of our total reward:
\begin{equation}
    R_{\text{accuracy}}(\mathcal{A}, B) = 
    \begin{cases}
        1 / \sqrt{N \cdot k} & \text{if } \exists p_i \in \mathcal{A} \text{ s.t. } p_i \in B \\
        -1 / N & \text{otherwise}
    \end{cases}
\end{equation}
This reward structure dynamically encourages wider exploration upon failure and rewards efficient, confident predictions upon success.

\paragraph{Collinear Penalty.}
\label{sec:collinear_penalty}
To further improve the quality of exploration, we introduce a penalty for low-quality exploration strategies. If the set of generated points $\mathcal{A}$ is found to be approximately collinear, we override the accuracy reward with a large negative value, $R_{\text{accuracy}} = -1$. Collinearity is determined by checking if the area of the triangle formed by any three points in the set is close to zero. This discourages the model from adopting trivial, inefficient linear scanning strategies and incentivizes more spatially diverse exploration.

\subsection{Overall Training Objective}
\label{sec:overall_objective}
The final reward signal for policy optimization combines a format reward $R_{\text{format}}$ with the accuracy reward $R_{\text{accuracy}}$. The format reward, which is +1 if the output string is correctly structured and 0 otherwise, serves as a prerequisite for any subsequent reward evaluation. The total reward is thus:
\begin{equation}
    R_{\text{total}} = R_{\text{format}} + R_{\text{accuracy}}
\end{equation}
This total reward is then used to compute an advantage estimate, $\hat{A}$, which directly guides the update of the policy parameters. The complete training process is outlined in Algorithm \ref{alg:aepo}.

\begin{algorithm}[!th]
\caption{AEPO Training Loop}
\label{alg:aepo}
\begin{algorithmic}[1]
\STATE Initialize model parameters $\theta$
\FOR{each training iteration}
    \STATE Sample $(\mathcal{S}, \mathcal{I}, B)$ from dataset $\mathcal{D}$
    \STATE Generate output sequence $\sigma \sim \pi_{\theta}(\cdot | \mathcal{S}, \mathcal{I})$
    \STATE $R_{\text{format}} \leftarrow \text{CheckFormat}(\sigma)$
    \STATE $R_{\text{accuracy}} \leftarrow 0$
    \IF{$R_{\text{format}} > 0$}
        \STATE Extract $N$ points $\mathcal{A} = \{p_1, ..., p_N\}$ from $\sigma$
        \IF{IsCollinear($\mathcal{A}$)}
            \STATE $R_{\text{accuracy}} \leftarrow -1$
        \ELSE
            \STATE $k \leftarrow \text{FindFirstCorrectRank}(\mathcal{A}, B)$
            \IF{$k$ is not None}
                \STATE $R_{\text{accuracy}} \leftarrow 1 / \sqrt{N \cdot k}$
            \ELSE
                \STATE $R_{\text{accuracy}} \leftarrow -1 / N$
            \ENDIF
        \ENDIF
    \ENDIF
    \STATE $R_{\text{total}} \leftarrow R_{\text{format}} + R_{\text{accuracy}}$
    \STATE Calculate advantage estimate $\hat{A}(\sigma, B)$ based on $R_{\text{total}}$
    \STATE Update $\theta$ using policy gradient with advantage $\hat{A}(\sigma, B)$
\ENDFOR
\end{algorithmic}
\end{algorithm}

\begin{table*}[!htp]
    \centering
    \small
    \caption{Performance comparison on the \textbf{MMBench-GUI} benchmark. We report top-1 accuracy (\%); for InfiGUI-G1 models, only the first generated answer is evaluated. Best and second-best results are shown in \textbf{bold} and \underline{underlined}, respectively. For our models, we also report the \textit{Exploration Success Rate} with the average number of generated candidates (\textit{Avg. N}), and standard deviation $\sigma$ over 5 runs.}
    % The table now has 15 columns: Model + 12 data cols + Avg + Std (σ)
    \begin{tabularx}{\textwidth}{l *{12}{>{\centering\arraybackslash}X} c c}
    \toprule
    % Main Headers updated to include a separate Std. dev. column
    \multirow{2}{*}{\textbf{Model}} & \multicolumn{2}{c}{\textbf{Windows}} & \multicolumn{2}{c}{\textbf{MacOS}} & \multicolumn{2}{c}{\textbf{Linux}} & \multicolumn{2}{c}{\textbf{iOS}} & \multicolumn{2}{c}{\textbf{Android}} & \multicolumn{2}{c}{\textbf{Web}} & \multirow{2}{*}{\textbf{Avg.}} & \multirow{2}{*}{\textbf{$\sigma$}} \\
    \cmidrule(lr){2-3} \cmidrule(lr){4-5} \cmidrule(lr){6-7} \cmidrule(lr){8-9} \cmidrule(lr){10-11} \cmidrule(lr){12-13}
    % Sub-headers for Basic and Advanced
    & Basic & Adv. & Basic & Adv. & Basic & Adv. & Basic & Adv. & Basic & Adv. & Basic & Adv. & & \\
    \midrule
    
    % --- Section 1: External Baselines ---
    % \multicolumn{2}{c}{...} is used to merge the Avg. and σ columns for models without std. dev.
    GPT-4o~\citep{hurst2024gpt} & 1.5 & 1.1 & 8.7 & 4.3 & 1.1 & 1.0 & 5.1 & 3.3 & 2.5 & 1.4 & 3.2 & 2.9 & \multicolumn{2}{c}{\cellcolor{lightgold!50}2.9} \\
    Claude-3.7~\citep{anthropic2024claude37} & 1.5 & 0.7 & 12.5 & 7.5 & 1.1 & 0.0 & 13.7 & 10.6 & 1.4 & 1.4 & 3.2 & 2.3 & \multicolumn{2}{c}{\cellcolor{lightgold!50}4.7} \\
    Qwen-Max-VL~\citep{bai2023qwen} & 43.9 & 36.8 & 58.8 & 56.1 & 53.9 & 30.1 & 77.4 & 59.1 & 79.5 & 70.1 & 74.8 & 58.8 & \multicolumn{2}{c}{\cellcolor{lightgold!50}58.0} \\
    ShowUI-2B~\citep{lin2024showui} & 9.2 & 4.4 & 24.1 & 10.4 & 25.1 & 11.7 & 29.0 & 19.7 & 17.4 & 8.7 & 22.9 & 12.7 & \multicolumn{2}{c}{\cellcolor{lightgold!50}16.0} \\
    Qwen2.5-VL-7B~\citep{bai2025qwen2} & 31.4 & 16.5 & 31.3 & 22.0 & 21.5 & 12.2 & 66.6 & 55.2 & 35.1 & 35.2 & 40.3 & 32.5 & \multicolumn{2}{c}{\cellcolor{lightgold!50}33.9} \\
    Qwen2.5-VL-72B~\citep{bai2025qwen2} & 55.7 & 33.8 & 49.9 & 30.1 & 40.3 & 20.9 & 56.1 & 28.2 & 55.6 & 25.4 & 68.4 & 45.8 & \multicolumn{2}{c}{\cellcolor{lightgold!50}41.8} \\
    OS-Atlas-Base-7B~\citep{wu2024osatlasfoundationactionmodel} & 36.9 & 18.8 & 44.4 & 21.7 & 31.4 & 13.3 & 74.8 & 48.8 & 69.6 & 46.8 & 61.3 & 35.4 & \multicolumn{2}{c}{\cellcolor{lightgold!50}41.4} \\
    Aguvis-7B-720P~\citep{xu2025aguvisunifiedpurevision} & 37.3 & 21.7 & 48.1 & 33.3 & 33.5 & 25.0 & 67.5 & 65.2 & 61.0 & 51.0 & 61.6 & 45.5 & \multicolumn{2}{c}{\cellcolor{lightgold!50}45.7} \\
    UI-TARS-1.5-7B~\citep{qin2025uitarspioneeringautomatedgui} & 68.3 & 39.0 & 69.0 & 44.5 & 64.4 & 37.8 & 88.5 & 69.4 & 90.5 & 69.3 & 81.0 & 56.5 & \multicolumn{2}{c}{\cellcolor{lightgold!50}64.3} \\
    UI-TARS-72B-DPO~\citep{qin2025uitarspioneeringautomatedgui} & 78.6 & 51.8 & 80.3 & \underline{62.7} & \underline{68.6} & \underline{51.5} & 90.8 & 81.2 & 93.0 & 80.0 & 88.1 & 68.5 & \multicolumn{2}{c}{\cellcolor{lightgold!50}74.3} \\
    UGround-V1-7B~\citep{gou2025navigatingdigitalworldhumans} & 66.8 & 39.0 & 71.3 & 48.6 & 56.5 & 31.1 & 92.7 & 70.9 & 93.5 & 71.0 & 88.7 & 64.6 & \multicolumn{2}{c}{\cellcolor{lightgold!50}65.7} \\
    InternVL3-72B~\citep{zhu2025internvl3exploringadvancedtraining} & 70.1 & 42.6 & 75.7 & 52.3 & 59.2 & 41.3 & 93.6 & 80.6 & 92.7 & 78.6 & 90.7 & 65.9 & \multicolumn{2}{c}{\cellcolor{lightgold!50}72.2} \\
    \midrule

    % --- Section 2: Internal Baselines ---
    Naive RLVR-3B & 68.6 & 44.5 & 78.6 & 50.0 & 61.3 & 39.3 & 92.4 & 76.4 & 91.3 & 76.1 & 87.4 & 63.0 & \multicolumn{2}{c}{\cellcolor{lightgold!50}70.9} \\
    Naive RLVR-7B & \underline{79.3} & \underline{58.1} & \underline{82.3} & \underline{62.7} & 64.4 & 44.9 & \underline{94.9} & \underline{89.1} & \pmb{95.5} & \underline{84.2} & \underline{92.9} & \pmb{79.5} & \multicolumn{2}{c}{\cellcolor{lightgold!50}\underline{79.3}} \\
    \midrule

    % --- Section 3: Ours ---
    \rowcolor{gray!15}
    \textbf{InfiGUI-G1-3B} & 74.2 & 47.1 & 78.8 & 55.2 & 65.4 & 41.8 & \pmb{95.2} & 78.8 & 92.1 & 78.0 & 89.7 & 64.3 & \cellcolor{lightgold}73.4 & \cellcolor{lightgold}\textit{\color{gray!80!black}{0.25}} \\
    \multicolumn{1}{r}{\textit{\textit{w/ Expl. Success \scriptsize\color{gray!80!black}{(Avg. N=2.0)}}}} & 79.7 & 59.9 & 86.4 & 66.8 & 73.3 & 54.1 & 97.1 & 87.0 & 96.3 & 88.7 & 95.2 & 75.6 & \cellcolor{lightgold!50}81.6 & \cellcolor{lightgold!50}\textit{\color{gray!80!black}{0.41}} \\
    \rowcolor{gray!15}
    \textbf{InfiGUI-G1-7B} & \pmb{82.7} & \pmb{61.8} & \pmb{83.8} & \pmb{63.9} & \pmb{72.3} & \pmb{52.0} & \underline{94.9} & \pmb{89.4} & \underline{95.2} & \pmb{85.6} & \pmb{93.5} & \underline{76.3} & \cellcolor{lightgold}\pmb{80.8} & \cellcolor{lightgold}\textit{\color{gray!80!black}{0.21}} \\
    \multicolumn{1}{r}{\textit{\textit{w/ Expl. Success \scriptsize\color{gray!80!black}{(Avg. N=1.6)}}}} & 87.1 & 69.1 & 87.2 & 76.3 & 78.5 & 58.2 & 98.1 & 92.4 & 98.0 & 91.8 & 97.1 & 85.7 & \cellcolor{lightgold!50}86.4 & \cellcolor{lightgold!50}\textit{\color{gray!80!black}{0.11}} \\
    \bottomrule
    \end{tabularx}
    \label{tab:main_results_mmbench}
\end{table*}

\begin{table*}[!htp]
    \centering
    \small
    \caption{Performance comparison on the \textbf{ScreenSpot-Pro} benchmark. We report Top-1 accuracy (\%); for multi-answer models, only the first generated answer is evaluated. Best and second-best scores are shown in \textbf{bold} and \underline{underlined}, respectively. For our models, we also report the \textit{Exploration Success Rate} with the average number of generated candidates (\textit{Avg. N}), and standard deviation $\sigma$ over 5 runs.}

    % The table now has 15 columns: Model + 12 data cols + Avg + Std (σ)
    \begin{tabularx}{\textwidth}{l *{12}{>{\centering\arraybackslash}X} c c}
    \toprule
    % Main Headers updated to include a separate Std. dev. column
    \multirow{2}{*}{\textbf{Model}} & \multicolumn{2}{c}{\textbf{CAD}} & \multicolumn{2}{c}{\textbf{Dev.}} & \multicolumn{2}{c}{\textbf{Creative}} & \multicolumn{2}{c}{\textbf{Scientific}} & \multicolumn{2}{c}{\textbf{Office}} & \multicolumn{2}{c}{\textbf{OS}} & \multirow{2}{*}{\textbf{Avg.}} & \multirow{2}{*}{\textbf{$\sigma$}} \\
    \cmidrule(lr){2-3} \cmidrule(lr){4-5} \cmidrule(lr){6-7} \cmidrule(lr){8-9} \cmidrule(lr){10-11} \cmidrule(lr){12-13}
    % Sub-headers for Text and Icon
    & Text & Icon & Text & Icon & Text & Icon & Text & Icon & Text & Icon & Text & Icon & & \\
    \midrule
    
    % --- Section 1: External Baselines ---
    GPT-4o~\citep{hurst2024gpt} & 2.0 & 0.0 & 1.3 & 0.0 & 1.0 & 0.0 & 2.1 & 0.0 & 1.1 & 0.0 & 0.0 & 0.0 & \multicolumn{2}{c}{\cellcolor{lightgold!50}0.8} \\
    Claude Comp. Use~\citep{anthropic2024b} & 14.5 & 3.7 & 22.0 & 3.9 & 25.9 & 3.4 & 33.9 & 15.8 & 30.1 & 16.3 & 11.0 & 4.5 & \multicolumn{2}{c}{\cellcolor{lightgold!50}17.1} \\
    SeeClick~\citep{cheng2024seeclick} & 2.5 & 0.0 & 0.6 & 0.0 & 1.0 & 0.0 & 3.5 & 0.0 & 1.1 & 0.0 & 2.8 & 0.0 & \multicolumn{2}{c}{\cellcolor{lightgold!50}1.1} \\
    Qwen2-VL-7B~\citep{wang2024qwen2} & 0.5 & 0.0 & 2.6 & 0.0 & 1.5 & 0.0 & 6.3 & 0.0 & 3.4 & 1.9 & 0.9 & 0.0 & \multicolumn{2}{c}{\cellcolor{lightgold!50}1.6} \\
    CogAgent-18B~\citep{hong2024cogagent} & 7.1 & 3.1 & 14.9 & 0.7 & 9.6 & 0.0 & 22.2 & 1.8 & 13.0 & 0.0 & 5.6 & 0.0 & \multicolumn{2}{c}{\cellcolor{lightgold!50}7.7} \\
    UI-R1-3B~\citep{lu2025ui} & 11.2 & 6.3 & 22.7 & 4.1 & 27.3 & 3.5 & 42.4 & 11.8 & 32.2 & 11.3 & 13.1 & 4.5 & \multicolumn{2}{c}{\cellcolor{lightgold!50}17.8} \\
    ZonUI-3B~\citep{hsieh2025zonui3blightweightvisionlanguagemodel} & 31.9 & 15.6 & 24.6 & 6.2 & 40.9 & 7.6 & 54.8 & 18.1 & 57.0 & 26.4 & 19.6 & 7.8 & \multicolumn{2}{c}{\cellcolor{lightgold!50}28.7} \\
    GUI-R1-7B~\citep{xia2025gui} & 23.9 & 6.3 & 49.4 & 4.8 & 38.9 & 8.4 & 55.6 & 11.8 & 58.7 & 26.4 & 42.1 & 16.9 & \multicolumn{2}{c}{\cellcolor{lightgold!50}31.0} \\
    UI-TARS-7B~\citep{qin2025ui} & 20.8 & 9.4 & 58.4 & 12.4 & 50.0 & 9.1 & 63.9 & \underline{31.8} & 63.3 & 20.8 & 30.8 & 16.9 & \multicolumn{2}{c}{\cellcolor{lightgold!50}35.7} \\
    UI-AGILE-7B~\citep{lian2025uiagileadvancingguiagents} & 49.2 & 14.1 & 64.3 & 15.2 & 53.0 & 9.8 & 72.9 & 25.5 & \underline{75.1} & 30.2 & 45.8 & 20.2 & \multicolumn{2}{c}{\cellcolor{lightgold!50}44.0} \\
    GUI-G$^2$-7B~\citep{tang2025guig2gaussianrewardmodeling} & \underline{55.8} & 12.5 & 68.8 & 17.2 & 57.1 & \underline{15.4} & \underline{77.1} & 24.5 & 74.0 & 32.7 & \pmb{57.9} & \underline{21.3} & \multicolumn{2}{c}{\cellcolor{lightgold!50}47.5} \\
    \midrule

    % --- Section 2: Internal Baselines ---
    Naive RLVR-3B & 36.0 & 18.8 & 63.0 & 15.2 & 49.5 & 13.3 & 65.3 & 26.4 & 64.4 & 32.1 & 39.3 & 16.9 & \multicolumn{2}{c}{\cellcolor{lightgold!50}39.8} \\
    Naive RLVR-7B & 53.8 & 17.2 & \underline{71.4} & 15.9 & \underline{60.6} & 11.9 & 76.4 & 26.4 & 74.6 & \underline{34.0} & 54.2 & 20.2 & \multicolumn{2}{c}{\cellcolor{lightgold!50}\underline{47.6}} \\
    \midrule

    % --- Section 3: Ours ---
    \rowcolor{gray!15}
    \textbf{InfiGUI-G1-3B} & 50.8 & \pmb{25.0} & 64.9 & \underline{20.0} & 51.5 & \pmb{16.8} & 68.8 & \pmb{32.7} & 70.6 & 32.1 & 49.5 & 15.7 & \cellcolor{lightgold}45.2 & \cellcolor{lightgold}\textit{\color{gray!80!black}{0.13}} \\
    \multicolumn{1}{r}{\textit{w/ Expl. Success \scriptsize\color{gray!80!black}{(Avg. N=2.1)}}} & 56.9 & 31.3 & 70.8 & 25.5 & 63.6 & 23.1 & 74.3 & 39.1 & 79.1 & 37.7 & 54.2 & 19.1 & \cellcolor{lightgold!50}52.0 & \cellcolor{lightgold!50}\textit{\color{gray!80!black}{0.17}} \\
    \rowcolor{gray!15}
    \textbf{InfiGUI-G1-7B} & \pmb{57.4} & \underline{23.4} & \pmb{74.7} & \pmb{24.1} & \pmb{64.6} & \underline{15.4} & \pmb{80.6} & \underline{31.8} & \pmb{75.7} & \pmb{39.6} & \underline{57.0} & \pmb{29.2} & \cellcolor{lightgold}\pmb{51.9} & \cellcolor{lightgold}\textit{\color{gray!80!black}{0.48}} \\
    \multicolumn{1}{r}{\textit{w/ Expl. Success \scriptsize\color{gray!80!black}{(Avg. N=2.0)}}} & 65.5 & 26.6 & 85.1 & 30.3 & 71.2 & 20.3 & 84.7 & 33.6 & 81.4 & 47.2 & 60.7 & 37.1 & \cellcolor{lightgold!50}58.0 & \cellcolor{lightgold!50}\textit{\color{gray!80!black}{0.24}} \\
    \bottomrule
    \end{tabularx}
    \label{tab:main_results_screenspot_pro}
\end{table*}
\begin{table*}[!htp]
    \centering
    \small
    \caption{Performance comparison on the \textbf{UI-Vision} benchmark. We report Top-1 accuracy (\%); For our models, only the first generated answer is evaluated. Best and second-best scores are shown in \textbf{bold} and \underline{underlined}, respectively. For our models, we also report the \textit{Exploration Success Rate} with the average number of generated candidates (\textit{Avg. N}), and standard deviation $\sigma$ over 5 runs.}
    % The table now has 13 columns: Model + 10 data cols + Overall + Std (σ)
    \begin{tabularx}{\textwidth}{l *{10}{>{\centering\arraybackslash}X} c c}
    \toprule
    % Header now includes both "Grouped by Category" and "Grouped by Setting"
    \multirow{2}{*}{\textbf{Model}} & \multicolumn{6}{c}{\textbf{Grouped by Category}} & \multicolumn{3}{c}{\textbf{Grouped by Setting}} & \multirow{2}{*}{\textbf{Overall}} & \multirow{2}{*}{\textbf{$\sigma$}} \\
    \cmidrule(lr){2-7} \cmidrule(lr){8-10}
    & Edu. & Browser & Dev. & Prod. & Creative & Entert. & Basic & Func. & Spatial & & \\
    \midrule
    
    % --- Section 1: External Baselines ---
    GPT-4o~\citep{hurst2024gpt} & 1.5 & 0.0 & 2.2 & 1.1 & 0.8 & 4.2 & 1.6 & 1.5 & 1.0 & \multicolumn{2}{c}{\cellcolor{lightgold!50}1.4} \\
    Claude-3.7-Sonnet~\citep{anthropic2024claude37} & 6.1 & 9.8 & 8.0 & 9.4 & 7.7 & 8.3 & 9.5 & 7.7 & 7.6 & \multicolumn{2}{c}{\cellcolor{lightgold!50}8.3} \\
    Qwen-2.5VL-7B~\citep{bai2025qwen2} & 0.5 & 0.0 & 1.2 & 0.9 & 0.5 & 1.0 & 1.2 & 0.8 & 0.5 & \multicolumn{2}{c}{\cellcolor{lightgold!50}0.9} \\
    InternVL2.5-8B~\citep{chen2025expandingperformanceboundariesopensource} & 1.1 & 7.0 & 3.0 & 1.8 & 1.2 & 5.2 & 2.5 & 2.8 & 1.0 & \multicolumn{2}{c}{\cellcolor{lightgold!50}2.1} \\
    MiniCPM-V-8B~\citep{yao2024minicpm} & 3.0 & 16.8 & 5.4 & 3.8 & 2.1 & 13.0 & 7.1 & 5.3 & 1.5 & \multicolumn{2}{c}{\cellcolor{lightgold!50}4.3} \\
    SeeClick-9.6B~\citep{cheng2024seeclick} & 4.2 & 13.3 & 7.3 & 4.3 & 4.0 & 11.0 & 9.4 & 4.7 & 2.1 & \multicolumn{2}{c}{\cellcolor{lightgold!50}5.4} \\
    ShowUI-2B~\citep{lin2024showui} & 3.7 & 13.3 & 7.5 & 6.5 & 2.5 & 15.6 & 8.1 & 7.7 & 2.1 & \multicolumn{2}{c}{\cellcolor{lightgold!50}5.9} \\
    CogAgent-9B~\citep{hong2024cogagentvisuallanguagemodel} & 8.7 & 11.2 & 8.6 & 10.3 & 5.6 & 15.6 & 12.0 & 12.2 & 2.6 & \multicolumn{2}{c}{\cellcolor{lightgold!50}8.9} \\
    OSAtlas-7B~\citep{wu2024osatlasfoundationactionmodel} & 8.7 & 16.8 & 10.3 & 9.2 & 5.6 & 16.2 & 12.2 & 11.2 & 3.7 & \multicolumn{2}{c}{\cellcolor{lightgold!50}9.0} \\
    AriaUI-25.3B~\citep{yang2025ariauivisualgroundinggui} & 9.0 & 18.9 & 11.2 & 10.4 & 6.5 & 19.3 & 12.2 & 14.0 & 4.0 & \multicolumn{2}{c}{\cellcolor{lightgold!50}10.1} \\
    UGround-v1-7B~\citep{gou2025navigatingdigitalworldhumans} & 10.4 & 28.7 & 17.5 & 12.2 & 8.6 & 18.2 & 15.4 & 17.1 & 6.3 & \multicolumn{2}{c}{\cellcolor{lightgold!50}12.9} \\
    UGround-v1-72B~\citep{gou2025navigatingdigitalworldhumans} & 22.4 & 35.7 & 27.6 & 21.6 & \pmb{18.3} & 38.0 & 27.9 & 26.7 & \pmb{14.9} & \multicolumn{2}{c}{\cellcolor{lightgold!50}23.2} \\
    Aguvis-7B~\citep{xu2025aguvisunifiedpurevision} & 13.1 & 30.8 & 17.1 & 12.1 & 9.6 & 24.0 & 17.8 & 18.3 & 5.1 & \multicolumn{2}{c}{\cellcolor{lightgold!50}13.7} \\
    UI-TARS-7B~\citep{qin2025uitarspioneeringautomatedgui} & 14.2 & 35.0 & 19.7 & 18.3 & 11.1 & 38.5 & 20.1 & 24.3 & 8.4 & \multicolumn{2}{c}{\cellcolor{lightgold!50}17.6} \\
    UI-TARS-72B~\citep{qin2025uitarspioneeringautomatedgui} & \underline{24.8} & 40.5 & \underline{27.9} & \pmb{26.8} & \underline{17.8} & 41.1 & 31.4 & 30.5 & \underline{14.7} & \multicolumn{2}{c}{\cellcolor{lightgold!50}\underline{25.5}} \\
    \midrule

    % --- Section 2: Internal Baselines ---
    Naive RLVR-3B & 18.5 & 37.8 & 21.8 & 19.6 & 12.8 & 42.7 & 27.4 & 24.6 & 7.3 & \multicolumn{2}{c}{\cellcolor{lightgold!50}19.4} \\
    Naive RLVR-7B & 23.5 & 42.7 & 27.4 & 24.5 & 16.2 & \underline{50.5} & \underline{32.9} & \underline{30.7} & 10.1 & \multicolumn{2}{c}{\cellcolor{lightgold!50}24.1} \\
    \midrule

    % --- Section 3: Ours ---
    \rowcolor{gray!15}
    \textbf{InfiGUI-G1-3B} & 22.6 & \underline{43.4} & 24.3 & 22.6 & 14.0 & 47.4 & 31.2 & 28.0 & 8.2 & \cellcolor{lightgold}22.0 & \cellcolor{lightgold}\textit{\color{gray!80!black}{0.20}} \\
    \multicolumn{1}{r}{\textit{w/ Expl. Success \scriptsize\color{gray!80!black}{(Avg. N=2.1)}}} & 29.3 & 51.7 & 30.5 & 31.7 & 20.5 & 59.9 & 39.2 & 36.7 & 14.6 & \cellcolor{lightgold!50}29.7 & \cellcolor{lightgold!50}\textit{\color{gray!80!black}{0.29}} \\
    \rowcolor{gray!15}
    \textbf{InfiGUI-G1-7B} & \pmb{25.5} & \pmb{46.2} & \pmb{29.6} & \underline{26.7} & 17.6 & \pmb{52.1} & \pmb{36.2} & \pmb{31.9} & 11.5 & \cellcolor{lightgold}\pmb{26.1} & \cellcolor{lightgold}\textit{\color{gray!80!black}{0.05}} \\
    \multicolumn{1}{r}{\textit{w/ Expl. Success \scriptsize\color{gray!80!black}{(Avg. N=2.1)}}} & 35.4 & 52.4 & 35.5 & 37.3 & 23.3 & 66.1 & 44.4 & 40.7 & 19.5 & \cellcolor{lightgold!50}34.4 & \cellcolor{lightgold!50}\textit{\color{gray!80!black}{0.12}} \\
    \bottomrule
    \end{tabularx}
    \label{tab:main_results_uivision}
\end{table*}
\begin{table*}[!htp]
    \centering
    \small
    \caption{Performance comparison on the \textbf{UI-I2E-Bench} benchmark. We report Top-1 accuracy (\%); For our models, only the first generated answer is evaluated. Best and second-best scores are shown in \textbf{bold} and \underline{underlined}, respectively. For our models, we also report the \textit{Exploration Success Rate} with the average number of generated candidates (\textit{Avg. N}), and standard deviation $\sigma$ over 5 runs.}
    % The table has 9 columns: Model + 3 Platform + 2 Implicitness + Overall + Std (σ)
    \begin{tabularx}{\textwidth}{l *{5}{>{\centering\arraybackslash}X} c c}
    \toprule
    % Main Headers
    \multirow{2}{*}{\textbf{Model}} & \multicolumn{3}{c}{\textbf{Grouped by Platform}} & \multicolumn{2}{c}{\textbf{Grouped by Implicitness}} & \multirow{2}{*}{\textbf{Overall}} & \multirow{2}{*}{\textbf{$\sigma$}} \\
    \cmidrule(lr){2-4} \cmidrule(lr){5-6}
    % Sub-headers
    & Web & Desktop & Mobile & Explicit & Implicit & & \\
    \midrule
    
    % --- Section 1: External Baselines ---
    Qwen2.5-VL-3B~\citep{bai2025qwen2} & 39.9 & 38.7 & 44.5 & 51.4 & 35.8 & \multicolumn{2}{c}{\cellcolor{lightgold!50}41.7} \\
    Qwen2.5-VL-7B~\citep{bai2025qwen2} & 56.9 & 41.6 & 61.7 & 58.4 & 51.0 & \multicolumn{2}{c}{\cellcolor{lightgold!50}53.8} \\
    Qwen2.5-VL-72B~\citep{bai2025qwen2} & 49.0 & 47.2 & 55.3 & 49.6 & 52.5 & \multicolumn{2}{c}{\cellcolor{lightgold!50}51.4} \\
    OS-Atlas-4B~\citep{wu2024osatlasfoundationactionmodel} & 54.6 & 19.9 & 58.6 & 51.5 & 39.9 & \multicolumn{2}{c}{\cellcolor{lightgold!50}44.3} \\
    OS-Atlas-7B~\citep{wu2024osatlasfoundationactionmodel} & 52.2 & 48.9 & 68.1 & 63.2 & 55.8 & \multicolumn{2}{c}{\cellcolor{lightgold!50}58.6} \\
    Aguvis-7B~\citep{xu2025aguvisunifiedpurevision} & 45.1 & 47.6 & 60.3 & 61.1 & 48.4 & \multicolumn{2}{c}{\cellcolor{lightgold!50}53.2} \\
    Uground-V1-2B~\citep{gou2025navigatingdigitalworldhumans} & 66.4 & 49.5 & 59.9 & 72.9 & 47.9 & \multicolumn{2}{c}{\cellcolor{lightgold!50}57.4} \\
    Uground-V1-7B~\citep{gou2025navigatingdigitalworldhumans} & 70.8 & 65.7 & 73.5 & 81.3 & 63.6 & \multicolumn{2}{c}{\cellcolor{lightgold!50}70.3} \\
    Uground-V1-72B~\citep{gou2025navigatingdigitalworldhumans} & 74.7 & \pmb{74.6} & 78.2 & 84.5 & \underline{71.3} & \multicolumn{2}{c}{\cellcolor{lightgold!50}\underline{76.3}} \\
    UI-TARS-2B~\citep{qin2025uitarspioneeringautomatedgui} & 62.2 & 54.0 & 66.7 & 74.1 & 54.5 & \multicolumn{2}{c}{\cellcolor{lightgold!50}62.0} \\
    UI-TARS-7B~\citep{qin2025uitarspioneeringautomatedgui} & 56.5 & 58.0 & 65.7 & 71.4 & 55.3 & \multicolumn{2}{c}{\cellcolor{lightgold!50}61.4} \\
    UI-TARS-1.5-7B~\citep{qin2025uitarspioneeringautomatedgui} & 79.5 & 68.8 & 74.1 & 81.3 & 68.2 & \multicolumn{2}{c}{\cellcolor{lightgold!50}73.2} \\
    UI-TARS-72B~\citep{qin2025uitarspioneeringautomatedgui} & 77.1 & \underline{69.8} & 75.5 & 80.9 & 69.4 & \multicolumn{2}{c}{\cellcolor{lightgold!50}73.7} \\
    UI-I2E-VLM-4B~\citep{liu2025ui} & 60.9 & 38.9 & 61.4 & 61.9 & 48.3 & \multicolumn{2}{c}{\cellcolor{lightgold!50}53.4} \\
    UI-I2E-VLM-7B~\citep{liu2025ui} & 62.1 & 64.0 & 76.2 & 72.0 & 67.9 & \multicolumn{2}{c}{\cellcolor{lightgold!50}69.5} \\
    UI-R1-E-3B~\citep{lu2025ui} & - & - & - & - & - & \multicolumn{2}{c}{\cellcolor{lightgold!50}69.1} \\
    \midrule

    % --- Section 2: Internal Baselines ---
    Naive RLVR-3B & 74.7 & 62.0 & 78.9 & 81.3 & 65.8 & \multicolumn{2}{c}{\cellcolor{lightgold!50}71.6} \\
    Naive RLVR-7B & \underline{83.0} & 63.0 & 77.6 & \underline{84.8} & 70.2 & \multicolumn{2}{c}{\cellcolor{lightgold!50}75.8} \\
    \midrule

    % --- Section 3: Ours ---
    \rowcolor{gray!15}
    \textbf{InfiGUI-G1-3B} & 79.8 & 60.7 & \underline{78.9} & 81.1 & 67.5 & \cellcolor{lightgold}72.6 & \cellcolor{lightgold}\textit{\color{gray!80!black}{0.30}} \\
    \multicolumn{1}{r}{\textit{w/ Expl. Success \scriptsize\color{gray!80!black}{(Avg. N=2.0)}}} & 89.3 & 73.0 & 87.7 & 88.8 & 79.2 & \cellcolor{lightgold!50}82.8 & \cellcolor{lightgold!50}\textit{\color{gray!80!black}{0.51}} \\
    \rowcolor{gray!15}
    \textbf{InfiGUI-G1-7B} & \pmb{84.6} & 66.3 & \pmb{83.0} & \pmb{85.0} & \pmb{72.7} & \cellcolor{lightgold}\pmb{77.4} & \cellcolor{lightgold}\textit{\color{gray!80!black}{0.40}} \\
    \multicolumn{1}{r}{\textit{w/ Expl. Success \scriptsize\color{gray!80!black}{(Avg. N=1.6)}}} & 87.4 & 71.7 & 89.8 & 87.3 & 80.4 & \cellcolor{lightgold!50}83.0 & \cellcolor{lightgold!50}\textit{\color{gray!80!black}{0.47}} \\
    \bottomrule
    \end{tabularx}
    \label{tab:main_results_ui_i2e_bench}
\end{table*}
\begin{table*}[!htp]
    \centering
    \small
    \caption{Performance comparison on the \textbf{ScreenSpot-V2} benchmark. We report Top-1 accuracy (\%); For our models, only the first generated answer is evaluated. Best and second-best scores are shown in \textbf{bold} and \underline{underlined}, respectively. For our models, we also report the \textit{Exploration Success Rate} with the average number of generated candidates (\textit{Avg. N}), and standard deviation $\sigma$ over 5 runs.}
    % The table now has 9 columns: Model + 6 data cols + Avg + Std (σ)
    \begin{tabularx}{\textwidth}{l *{6}{>{\centering\arraybackslash}X} c c}
    \toprule
    % Main Headers for ScreenSpot-v2 benchmark
    \multirow{2}{*}{\textbf{Model}} & \multicolumn{2}{c}{\textbf{Mobile}} & \multicolumn{2}{c}{\textbf{Desktop}} & \multicolumn{2}{c}{\textbf{Web}} & \multirow{2}{*}{\textbf{Avg.}} & \multirow{2}{*}{\textbf{$\sigma$}} \\
    \cmidrule(lr){2-3} \cmidrule(lr){4-5} \cmidrule(lr){6-7}
    % Sub-headers for Text and Icon/Widget
    & Text & Icon/Widget & Text & Icon/Widget & Text & Icon/Widget & & \\
    \midrule
    
    % --- Section 1: External Baselines ---
    SeeClick~\citep{cheng2024seeclick} & 78.4 & 50.7 & 70.1 & 29.3 & 55.2 & 32.5 & \multicolumn{2}{c}{\cellcolor{lightgold!50}55.1} \\
    OS-Atlas-Base-7B~\citep{wu2024osatlasfoundationactionmodel} & 95.2 & 75.8 & 90.7 & 63.6 & 90.6 & 77.3 & \multicolumn{2}{c}{\cellcolor{lightgold!50}85.1} \\
    UI-TARS-7B~\citep{qin2025uitarspioneeringautomatedgui} & 96.9 & 89.1 & \underline{95.4} & \underline{85.0} & 93.6 & 85.2 & \multicolumn{2}{c}{\cellcolor{lightgold!50}91.6} \\
    UI-TARS-72B~\citep{qin2025uitarspioneeringautomatedgui} & 94.8 & 86.3 & 91.2 & \pmb{87.9} & 91.5 & \underline{87.7} & \multicolumn{2}{c}{\cellcolor{lightgold!50}90.3} \\
    Qwen2.5-VL-3B~\citep{bai2025qwen2} & 93.4 & 73.5 & 88.1 & 58.6 & 88.0 & 71.4 & \multicolumn{2}{c}{\cellcolor{lightgold!50}80.9} \\
    Qwen2.5-VL-7B~\citep{bai2025qwen2} & 97.6 & 87.2 & 90.2 & 74.2 & 93.2 & 81.3 & \multicolumn{2}{c}{\cellcolor{lightgold!50}88.8} \\
    Qwen2.5-VL-32B~\citep{bai2025qwen2} & 97.9 & 88.2 & \pmb{98.5} & 79.3 & 91.2 & 86.2 & \multicolumn{2}{c}{\cellcolor{lightgold!50}91.3} \\
    \midrule

    % --- Section 2: Internal Baselines ---
    Naive RLVR-3B & \pmb{99.3} & 86.3 & 93.3 & 80.7 & 94.0 & 79.8 & \multicolumn{2}{c}{\cellcolor{lightgold!50}90.1} \\
    Naive RLVR-7B & 99.0 & \underline{91.5} & 94.8 & 80.7 & \underline{96.6} & 85.2 & \multicolumn{2}{c}{\cellcolor{lightgold!50}\underline{92.5}} \\
    \midrule

    % --- Section 3: Ours ---
    \rowcolor{gray!15}
    \textbf{InfiGUI-G1-3B} & \pmb{99.3} & 88.2 & 94.8 & 82.9 & 94.9 & 80.3 & \cellcolor{lightgold}91.1 & \cellcolor{lightgold}\textit{\color{gray!80!black}{0.05}} \\
    \multicolumn{1}{r}{\textit{w/ Expl. Success \scriptsize\color{gray!80!black}{(Avg. N=2.0)}}} & 99.7 & 91.9 & 95.9 & 88.6 & 97.4 & 88.7 & \cellcolor{lightgold!50}94.4 & \cellcolor{lightgold!50}\textit{\color{gray!80!black}{0.12}} \\
    \rowcolor{gray!15}
    \textbf{InfiGUI-G1-7B} & 99.0 & \pmb{91.9} & 94.3 & 82.1 & \pmb{97.9} & \pmb{89.2} & \cellcolor{lightgold}\pmb{93.5} & \cellcolor{lightgold}\textit{\color{gray!80!black}{0.09}} \\
    \multicolumn{1}{r}{\textit{w/ Expl. Success \scriptsize\color{gray!80!black}{(Avg. N=1.4)}}} & 99.3 & 95.3 & 95.4 & 87.9 & 98.7 & 92.6 & \cellcolor{lightgold!50}95.6 & \cellcolor{lightgold!50}\textit{\color{gray!80!black}{0.12}} \\
    \bottomrule
    \end{tabularx}
    \label{tab:main_results_screenspot_v2}
\end{table*}

\section{Experiments}
\label{sec:experiments}

\subsection{Experimental Setup}
\label{sec:exp_setup}

\paragraph{Benchmarks and Metrics.}
We evaluate all models on five challenging benchmarks, each chosen to assess distinct capabilities. 
\textbf{MMBench-GUI}~\cite{wang2025mmbenchgui} is a comprehensive benchmark with a hierarchical design of basic and advanced instructions, which we use to evaluate the overall effectiveness of our method across tasks of varying complexity. 
\textbf{ScreenSpot-Pro}~\cite{li2025screenspotpro} is a benchmark designed to evaluate performance on high-resolution screens from professional software. Its distinct separation of text-based and icon-based grounding tasks provides a valuable setting to probe a model's semantic understanding, as icon grounding in particular requires associating abstract symbols with their functions.
\textbf{UI-Vision}~\cite{nayak2025uivisiondesktopcentricguibenchmark} is designed to test generalization across a wide variety of desktop applications, assessing the model's robustness in diverse, unseen environments.
Additionally, we report results on the widely-used \textbf{ScreenSpot-v2}~\cite{cheng2024seeclick,wu2024osatlasfoundationactionmodel} benchmark, which provides comprehensive coverage across mobile, desktop, and web platforms with a focus on both text and icon/widget elements. To further probe the semantic reasoning capabilities of the models, we also evaluate on \textbf{UI-I2E-Bench}~\cite{liu2025ui}. This next-generation benchmark was designed to overcome limitations of earlier datasets by including a higher proportion of implicit instructions that require semantic and spatial reasoning beyond direct text matching.
Our primary evaluation metric is \textbf{Accuracy}, where a prediction is considered correct if its coordinate point falls within the ground truth bounding box. For methods that output a bounding box, its center point is used. To demonstrate the high success rate of our exploration strategy, we also report the \textbf{Exploration Success Rate} for our InfiGUI-G1 models, where a sample is marked as a success if at least one of the generated candidate points is correct.

\paragraph{Baselines.}
To ensure a fair and rigorous comparison, we establish two sets of baselines. First, for controlled analysis, we train a Naive RLVR model for both size as \textbf{internal baselines}. It is trained using the exact same dataset and optimized hyperparameters as our core models. Second, to position our work within the broader literature, we compare it against several state-of-the-art models from recent works.

\paragraph{Implementation Details.}
Our InfiGUI-G1 models are built upon the open-source \textbf{Qwen2.5-VL-3B-Instruct} and \textbf{Qwen2.5-VL-7B-Instruct} as backbones. For the RLVR training phase, we adopt the RLOO algorithm~\citep{ahmadian2024back}, which effectively reduces the variance of policy gradient estimates by employing the average reward of other samples within the same batch as a baseline. This ``leave-one-out" strategy obviates the need for training a separate critic model. The RLOO policy gradient $\nabla_\theta J(\theta)$ is estimated as:
\begin{equation*}
    \begin{split}
        \nabla_\theta J(\theta) \approx &\frac{1}{k} \sum_{i=1}^{k} \left[ R(y_{(i)}, x) - \frac{1}{k-1} \sum_{j \neq i} R(y_{(j)}, x) \right] \\
        &\cdot \nabla_\theta \log \pi_\theta(y_{(i)}|x)
    \end{split}
\end{equation*}
where $k$ is the number of output sequences $y_{(i)}$ sampled from the policy $\pi_\theta$ given input $x$. Across all experiments, we employ a reasoning prompting paradigm, instructing the model to generate its reasoning process within \texttt{<think> </think>} tags before providing the final answer.

\paragraph{Training Details.}
Our training data is a mixture sampled from several public GUI datasets, including Widget Caption, OmniAct, GUICourse, etc., resulting in approximately 44k samples. Following common practices in RLVR to focus training on more challenging instances, we apply a data filtering strategy: for each sample, we generate 8 responses with a temperature of 1.0; if all 8 are correct, the sample is deemed too easy and is excluded. All models were trained on 16 H800 GPUs. Key training parameters include a learning rate of 1e-6, a rollout batch size of 128, and an RLOO rollout number of $n=8$. We train for 3 epochs.

\subsection{Main Results}
\label{sec:main_results}

We present the main results of our evaluation in Table~\ref{tab:main_results_mmbench}, \ref{tab:main_results_screenspot_pro}, \ref{tab:main_results_uivision}, \ref{tab:main_results_ui_i2e_bench}, and \ref{tab:main_results_screenspot_v2}. The results consistently show that our InfiGUI-G1 models establish new state-of-the-art performance among open-source models in both the 3B and 7B parameter categories. Notably, our models also exhibit competitive or superior performance against several proprietary models with significantly larger parameter counts, highlighting the efficacy and efficiency of our proposed AEPO framework.

The comparison with our internal baselines reveals that InfiGUI-G1 consistently and substantially outperforms the Naive RLVR model across all benchmarks. This direct comparison suggests that the performance gains can be attributed to the architectural and methodological improvements introduced by AEPO. Furthermore, our models demonstrate strong performance against other SOTA methods, including those based on SFT (e.g., UGround, OS-Atlas), many of which require training data exceeding 1M samples. In contrast, our approach achieves these competitive results using 44k instances, underscoring its data efficiency. Our results also show strong performance against other RLVR approaches that utilize IoU or distance-based rewards (e.g., GUI-R1, GUI-G$^2$).

Our method demonstrates strong generalization capabilities by achieving consistently high performance across multiple benchmarks with distinct focuses (e.g., UI-Vision, ScreenSpot-Pro). Crucially, these benchmarks contain many applications and scenarios not present in our training data, indicating that AEPO fosters a robust understanding rather than overfitting. The benefits of AEPO in enhancing semantic understanding appear particularly pronounced on the ScreenSpot-Pro benchmark. Here, our models show a more substantial improvement on icon-based grounding tasks than on text-based ones when compared to the Naive RLVR baseline, suggesting that AEPO's enhanced exploration is especially beneficial for tasks requiring association of abstract visual symbols with their functions.

\subsection{Ablation Studies}
\label{sec:ablation}

To dissect the contribution of each component within our AEPO framework, we conduct a series of ablation studies on the ScreenSpot-Pro benchmark. As its icon-based grounding tasks directly probe semantic understanding, this benchmark provides a clear setting to evaluate our design choices. The results are summarized in Table~\ref{tab:ablation_studies}.

\begin{table}[t]
    \centering
    \small
    \caption{Ablation study on the \textbf{ScreenSpot-Pro} benchmark. We compare model variants by Accuracy (\%), \textit{Exploration Success Rate} (\%), and average number of answers per sample. Best results within each group are shown in \textbf{bold}.}

    \label{tab:ablation_studies}
    \begin{tabular}{@{}l c c c@{}}
        \toprule
        \textbf{Model Configuration} & Acc. & Expl. Succ. & \# Answers \\
        \midrule
        \multicolumn{4}{c}{\textit{3B Models}} \\
        \midrule
        \rowcolor{gray!15}
        \textbf{InfiGUI-G1 (Full Model)} & \textbf{45.2} & \textbf{52.0} & 2.1 \\
        w/o Multi-Answer (Naive) & 40.4 & - & 1.0 \\
        w/o AER (use naive reward) & 38.4 & 42.1 & 1.9 \\
        w/o AER's rank factor \textit{k} & 38.1 & 47.6 & 2.5 \\
        w/o Collinear Penalty & 35.3 & 44.1 & 6.6 \\
        \midrule
        \multicolumn{4}{c}{\textit{7B Models}} \\
        \midrule
        \rowcolor{gray!15}
        \textbf{InfiGUI-G1 (Full Model)} & \textbf{51.9} & \textbf{58.0} & 2.0 \\
        w/o Multi-Answer (Naive) & 46.5 & - & 1.0 \\
        w/o AER (use naive reward) & 41.4 & 45.5 & 1.9 \\
        w/o AER's rank factor \textit{k} & 44.0 & 50.5 & 1.9 \\
        w/o Collinear Penalty & 37.0 & 43.8 & 8.2 \\
        \bottomrule
    \end{tabular}
\end{table}

The results reveal a clear logic behind AEPO's design. Removing \textbf{multi-answer generation} (`w/o Multi-Answer') leads to a significant performance drop, confirming that enabling exploration is the necessary first step. However, this exploration must be guided effectively, as replacing our \textbf{AER} with a naive reward (`w/o AER') causes a further decline. The importance of AER's ranking factor \textit{k} is particularly insightful; removing it (`w/o k') results in a model that often finds the correct answer (high Expl. Succ.) but fails to rank it first (low Acc.), demonstrating that \textit{k} is crucial for teaching the model \textbf{confidence} in its correct discoveries. Finally, the \textbf{collinear penalty} proves essential for ensuring the \textbf{quality} of exploration. Without it, the model adopts a degenerate strategy of generating numerous low-quality answers (high \# of answers) while accuracy plummets, showing the penalty is critical for preventing reward hacking.

\subsection{Analysis of AEPO's Effectiveness}
\label{sec:analysis}

To further understand the mechanisms of AEPO, we conduct three targeted analyses.

\paragraph{Adaptive Exploration Strategy.} We investigate if the model learns an adaptive exploration strategy. A clear correlation emerges between benchmark difficulty (indicated by model accuracy) and exploratory behavior. Our 7B model generates the most answers on the hardest benchmark (UI-Vision: 26.1\% Acc, 2.1 answers) and the fewest on the easiest (ScreenSpot-V2: 93.5\% Acc, 1.4 answers). This suggests AEPO learns to adaptively allocate exploratory resources based on task complexity.

\paragraph{Exploration Efficiency.}
We then evaluate the quality and efficiency of AEPO's exploration. Our InfiGUI-G1 models on ScreenSpot-Pro generate approximately two candidate answers per instance on average. To contextualize this, we compare our single-pass Exploration Success Rate against the multi-pass `pass@k` accuracy of the Naive RLVR baseline. As detailed in Table~\ref{tab:exploration_efficiency}, the results are compelling. Even when the Naive RLVR model is allowed four independent attempts (`pass@4`), its success rate in finding a correct answer is still significantly lower than that of our InfiGUI-G1, which achieves a higher success rate in a single pass with only about two attempts. This demonstrates that AEPO's multi-answer generation is not merely about increasing the number of tries, but about performing a more structured and efficient exploration of the action space.

\begin{table}[t]
\centering
\small
\caption{Exploration efficiency (\%) on ScreenSpot-Pro. Our single-pass success rate surpasses the baseline's multi-pass rate.}
\label{tab:exploration_efficiency}
\begin{tabular}{@{}lcc@{}}
\toprule
\textbf{Method} & \textbf{3B Models} & \textbf{7B Models} \\ \midrule
Naive RLVR (pass@2) & 41.7 & 49.8 \\
Naive RLVR (pass@4) & 43.5 & 52.1 \\
\rowcolor{gray!15}
\textbf{InfiGUI-G1 (Expl. Succ.)} & \textbf{52.0} & \textbf{58.0} \\ 
\multicolumn{1}{l}{\hspace{1em}$\hookrightarrow$ \textit{Avg. N}} & 2.1 & 2.0 \\
\bottomrule
\end{tabular}
\end{table}

\paragraph{Performance on Hard-to-Explore Samples.}
Finally, to validate our core hypothesis that AEPO resolves the exploration bottleneck, we designed an experiment to analyze performance on samples of varying difficulty. We partitioned the ScreenSpot-Pro test set by first using the base MLLM to generate 16 stochastic responses for each sample. Samples were then labeled as `hard` if the base model failed all 16 times, `easy` if it succeeded every time, and `middle` otherwise. The `hard` subset therefore represents samples that are highly unlikely to be answered correctly through naive exploration. As shown in Table~\ref{tab:hard_samples_analysis}, we then compared InfiGUI-G1 against the Naive RLVR baseline on these subsets. While our model improves performance across the board, the most significant gains are concentrated on the `hard` subset. On these critical samples, our 7B model achieves a relative improvement of over 60\%. This provides direct evidence that AEPO effectively creates learning signals for previously "unlearnable" samples, addressing the fundamental limitation we set out to solve.

\begin{table}[t]
\centering
\small
\caption{Accuracy (\%) on ScreenSpot-Pro subsets of varying difficulty. AEPO's advantage is most significant on `hard' samples.}
\label{tab:hard_samples_analysis}
\begin{tabular}{@{}l cc cc@{}}
\toprule
\multirow{2}{*}{\shortstack{\textbf{Difficulty}\\\textbf{Subset}}} & \multicolumn{2}{c}{\textbf{3B Models}} & \multicolumn{2}{c}{\textbf{7B Models}} \\
\cmidrule(lr){2-3} \cmidrule(lr){4-5}
& \shortstack{Naive\\RLVR} & \shortstack{InfiGUI-G1\\(Ours)} & \shortstack{Naive\\RLVR} & \shortstack{InfiGUI-G1\\(Ours)} \\ \midrule
Easy & 100 & \textbf{100} & 100 & \textbf{100} \\
Middle & 75.9 & \textbf{78.9} \textit{\scriptsize\color{gray!80!black}{(+4.0\%)}} & 72.6 & \textbf{78.4} \textit{\scriptsize\color{gray!80!black}{(+8.0\%)}} \\
Hard & 25.5 & \textbf{31.4} \textit{\scriptsize\color{gray!80!black}{(+23.1\%)}} & 10.8 & \textbf{17.4} \textit{\scriptsize\color{gray!80!black}{(+61.1\%)}} \\ \bottomrule
\end{tabular}
\end{table}
\section{Conclusion}
\label{sec:conclusion}

In this work, we addressed the critical challenge of enhancing semantic alignment in MLLM-based GUI agents, identifying the inefficient exploration of standard RLVR as a key bottleneck. We proposed AEPO, a policy optimization framework that integrates multi-answer generation with a theoretically-grounded AER function to enable effective exploration. Our model, InfiGUI-G1, achieves state-of-the-art performance, and our comprehensive analyses confirm that its effectiveness stems from its ability to adapt its exploration strategy, its high efficiency compared to naive sampling, and its success in creating learning signals for previously ``unlearnable" samples.

Limitations of our work include the computational overhead from multi-answer generation and a performance ceiling imposed by the backbone MLLM's visual capabilities, which could be addressed in future work by exploring more efficient sampling strategies and integration with more advanced visual encoders.

\bibliography{main}
\end{document}